\documentclass[runningheads]{llncs}
\usepackage{graphicx}
\usepackage{natbib}
\usepackage{amsmath}
\usepackage{graphicx}
\usepackage{amssymb}
\newcommand{\vecb}[1]{\boldsymbol{#1}}
\newcommand{\mat}[1]{\boldsymbol{\mathrm{#1}}}

\begin{document}
\title{Suggesting Cooking Recipes Through Simulation and Bayesian Optimization}
%
%
\author{Eduardo C. Garrido-Merch\'an \and
Alejandro Albarca-Molina}
\institute{Universidad Aut\'onoma de Madrid, Francisco Tom\'as y Valiente 11, Madrid, Spain
\email{eduardo.garrido@uam.es, alejandro.albarca@estudiante.uam.es}}

\maketitle              
\begin{abstract}
Cooking typically involves a plethora of decisions about ingredients and tools
that need to be chosen in order to write a good cooking recipe. Cooking can be modelled in an optimization framework, 
as it involves a search space of ingredients, kitchen tools, cooking times or temperatures. If we model as an objective 
function the quality of the recipe, several problems arise. No analytical expression can model all the recipes, so no gradients are available. The objective function is subjective, in other words, it contains noise. Moreover, evaluations are expensive 
both in time and human resources. Bayesian Optimization (BO) emerges as an ideal methodology to tackle problems with these characteristics. In this paper,
we propose a methodology to suggest recipe recommendations based on a Machine Learning (ML) model that fits real and simulated data and BO. We provide empirical evidence with two experiments that support the adequacy of the methodology.

\end{abstract}
\section{Introduction}

Creating good quality cooking recipes is an ancient art that has been developed by a wide variety of cultures. Nowadays, technology and modelling bring us new tools to experiment and test cooking recipes. Some solutions that have been studied are synthesizing cooking recipes using video \citep{doman2011video} or recommend cooking recipes based on preferences of a user \citep{ueda2011user}. But a few solutions tackle the problem of generating an optimal cooking recipe based on an input space of ingredients, tools and other criteria.

Lots of variables and decisions are involved in the process. From which ingredients to use and in which quantity to what tools to use, cooking times or intensity of the home appliances. We may use an optimization framework to model this problem. Let all these parameters lie in a space represented by $\mathbf{\Theta}$, where $\mathbf{\theta} \in \mathbf{\Theta}$ is a single configuration of that parameter space. We can model this space by establishing the parameters that are going to be used and the ranges that they fit in. For example, for a pasta dish, we may consider the boiling time minutes, $[5,15]$, and the quantity of used pasta grams $[100,300]$. $\theta = [8,150] \in \Theta$ would be a valid recipe that considers 8 boiling minutes and $150$ pasta grams.

All these parameters generate a cooking recipe that has to be evaluated. This evaluation is subjective. Let the quality of a recipe be represented by a scalar variable $\lambda \in [0,10]$, with $0$ being low quality and $10$ being high quality. If two or more individuals are given a recipe, their evaluation may differ. More formally, let $\mathbf{\lambda} = \mathbf{e}(\theta)$ be a vector of evaluators, where each $\lambda_i = f_i(\theta)$ represents the evaluation of a single evaluator and $\theta$ is the configuration of a recipe. Then, every $\lambda_i$ may be different w.r.t another one contained in $\mathbf{\lambda}$. We have a latent function $f$ contaminated by noise, $y(\theta) = f(\theta) + \epsilon : \epsilon \sim \mathcal{N}(\mu, \sigma)$. Other characteristics of the cooking process are that the evaluation is very expensive both in time, human and other resources. In this setting, we cannot depend on an optimization framework that needs lots of iterations to deliver a good solution to the problem, like metaheuristics. Moreover, in this problem, we do not have access to gradients as there is no analytical expression that models the cooking process. Hence, we can not use classical optimization techniques that find an optimal solution using gradients.

Bayesian Optimization (BO), described in a friendly way in this tutorial \citep{brochu2010tutorial}, emerges as a useful solution for problems with the described characteristics. BO relies on modelling the response surface by a probabilistic model, typically a Gaussian Processes (GP) \citep{rasmussen2004gaussian}, generating a surrogate model that is cheap to evaluate and gives predictions for all the input space that consider uncertainty. From this GP, BO builds an acquisition function that represents the utility of evaluating each point in order to find the optimum to the problem. 

In a real case scenario, where evaluations are performed while the optimization process is being done, BO is an adequate solution. But by only using BO, we need to cook as many times as iterations we need to perform, and for generating prototypes of recipes, that may be expensive in infrastructure and human resources. In this paper, we proposed a methodology to overcome this issue by not cooking for every iteration of the BO process, we will only need to cook before the process starts in order to build a Dataset.

We propose a methodology for generating prototype cooking recipes where we have simulated the cooking process by a fitted ML model ,$m$, that has approximated, $y_{approx}$, the cooking evaluation, $y$, by having learned it from data $\mathcal{D}$. This data $\mathcal{D}$ is the union of real data $\mathcal{D}_{real}$ and synthetic data $\mathcal{D}_{sim}$ generated by simulation. We have used these approximations to the evaluation and the data as we have the restriction of not having the human and infrastructure resources for evaluating each cooking recipe nor cooking thousands of meals. On the other hand, we have used BO since our objective is to corroborate the hypothesis that using BO, \citep{golovin2017google}, for this kind of problem is a good solution and that it also serves to optimize the cooking process if it is approximated by a fitted ML model, $m$.

The rest of the paper is divided as follows: Section 2 describes related work in this topic. Section 3 is an overview of BO basics. Section 4 presents the methodology that we have proposed to tackle this problem. Then, an experiments section provides empirical evidence that supports the adequacy of the proposed methodology. Finally, we sum up with Conclusions and Further Work around this topic.

\section{Related Work}

There is little work in this area, but some systems have been proposed to give a solution to this problem. The first one is IBM Chef Watson: \citep{lohr2013and}, which relies on the IBM Watson system that won the Jeopardy Contest. This system process natural language texts and uses predictive modelling to provide cooking recipes. It has the disadvantages that it relies on this private platform and it needs lots of data to fit the preferences of the user. Google Vizier, \citep{golovin2017google}, is a free platform for BO proposed by Google. They have used this tool to provide a recipe for chocolate cookies by having them tested by employees each day in their offices. Our approach differs in that we do not need to wait for several weeks to suggest a good quality cooking recipe. With our approach, we can provide a good quality cooking recipe in a single day if we have enough human resources to evaluate a set of meals.

\section{Black-Box Bayesian Optimization}

BO has been used in a plethora of scenarios: From Hyperparameter Tuning of ML Algorithms \citep{snoek2012practical} \citep{cordoba2018bayesian} to renewable energies \citep{cornejo2018bayesian} and a wide variety of applications \citep{shahriari2016taking}. BO process follows an iterative scheme where it uses a probabilistic model as a surrogate model, typically a Gaussian Process (GP) which is a prior over functions. This model serves as a prior for the objective function that we are optimizing, that is, we assume that the function can be sampled from a GP. 

The GP \citep{rasmussen2004gaussian} is a non-parametric model defined by a mean vector $\mathbf{\mu}$ and a covariance function $K(\mathbf{x}, \mathbf{x}')$ that defines a covariance matrix $\mathbf{K}$ , $f(\mathbf{x}) \approx GP(\mathbf{\mu}, \mathbf{K})$. By using BO we are assuming that the objective function $o(\mathbf{x})$ that we want to optimize can be sampled from the GP, which is going to be fitted by the evaluations $\mathcal{D}\{\mathbf{X},\mathbf{y}\}$ that we will be performing in the iterative scheme. The predictive distribution of the GP defines a mean $\mu_i$ and a standard deviation $\sigma_i$ for every point $\mathbf{x}$ where the objective function can be evaluated $y_i = o(\mathbf{x}) \approx N(\mu_i, \sigma_i)$, in other words, we have an uncertain prediction $N(\mu_i, \sigma_i)$ of every point of the space $\mathbf{x}$ where the objective function can be evaluated $y_i = o(\mathbf{x})$. The equations for the predictive distribution of the GP are:

\begin{equation}\label{eq:gp}
\begin{aligned}
        \mu & = \vecb{k}_{*}^{T}
        (\mat{K}+\sigma_{n}^{2}\mat{I})^{-1}\vecb{y}\,, \\
\sigma^2 & = k(\vecb{\theta}_t,\vecb{\theta}_t) -
        \boldsymbol{k}_{*}^T(\mat{K}+\sigma_{n}^{2}\mat{I})^{-1}\vecb{k}_*\,.
\end{aligned}
\end{equation}
where $\vecb{y} = (y_1, \ldots, y_{t-1})^t$ is a vector with the objective values observed so
far; $\sigma_n^2$ is the variance of the additive Gaussian noise; $\vecb{k}_*$ is a vector with the prior covariances between
$f(\vecb{\theta}_t)$ and each $y_i$; $\mat{K}$ is a matrix with the prior
covariances among each $y_i$; and
$k(\vecb{\theta}_t,\vecb{\theta}_t)$ is the prior variance at the candidate
location $\vecb{\theta}_t$.  {The} covariance
function $k(\cdot,\cdot)$ is pre-specified; for further
details about GPs and example of covariance functions we refer the reader to
\cite{rasmussen2004gaussian}.

In every iteration, the BO algorithm fits the GP with the point recommended from BO in the previous step, $\mathbf{x}^*$. This recommendation is obtained by the construction of an acquisition function, $\alpha(\mathcal{X})$, which is built from the predictive distribution $p(\mathcal{X})$ in every point of the input space $p(\mathbf{x})$ of the GP $\alpha(p(\mathbf{x}))$. The suggestion for the next evaluation is given by the maximum of this function $\mathbf{x}^* = \arg\max(\alpha(\mathcal{X}))$, where $\mathcal{X} \in \mathbb{R}^d$ is the bounded $d$-dimensional input space. There exist different criteria for this acquisition function $\alpha(\cdot)$ that basically tries to obtain a balanced tradeoff between exploitation of promising surfaces of the input space and exploration of unknown surfaces. Some examples of acquisition functions are the Expected Improvement (EI) \citep{brochu2010tutorial} or the Predictive Entropy Search (PES) \citep{hernandez2014predictive}. The key facts from BO are that the process of optimizing this acquisition function is cheap, it is typically done by extracting the best point from a grid and then starting there a local search algorithm such as the L-BFGS algorithm.

Once $\mathbf{x}^*$ is computed, it is added to the previous datasets of observations $\mathcal{D}\{\mathbf{X},\mathbf{y}\}$, the GP is conditioned over that point and the cycle starts again, building another acquisition function and extracting the maximum. This process is repeated until a convergence criterion is satisfied, typically when the budget of evaluations is satisfied. The final suggestion given by BO can be extracted by the best observation made so far. More details of BO can be found in this tutorial \citep{brochu2010tutorial}.

\section{Proposed Methodology: Simulation and Bayesian Optimization}

In a real case scenario, a possible methodology is to cook a recipe with a certain configuration $\mathbf{\theta}$, evaluate it by a jury of experts $y = f(\mathbf{\theta})$ and give that evaluation to BO, which will provide another suggestion to be cooked. As this is a very tedious, but possible scenario for restaurants, process and because our objective is to provide a prototype of an optimal recipe $\mathbf{\theta}^*$, we will approximate the evaluation function $f(\mathbf{\theta})$ given by experts by the prediction of an ML algorithm $m(\mathbf{\theta}) \approx f(\mathbf{\theta})$, which will be fitted by the union of real $\mathcal{D}_{real}$ and simulated data $\mathcal{D}_{sim}$.

In order to obtain prototypes of cooking recipes $\mathbf{\theta} \in \mathbf{\Theta}$, we need to follow the following steps. First of all, we need to obtain to define the bounded input space of parameters that are going to be involved in the cooking recipe $\mathbf{\Theta} \in \mathbb{R}^d$. Example of parameters $\mathbf{\theta} \in \mathbf{\Theta}$ could be the temperature of the oven or the meat grams used. We need to specify the lower $l_i$ and upper $u_i$ limits of each variable. In the BO process, GPs assume real variables but in the kitchen, we can find integer variables, intensity of the glass-ceramic or categorical variables, an ingredient brand. In order to overcome this issue, we use the approach given by Garrido \citep{garrido2018dealing}.

To build a dataset of real data $\mathcal{D}_{real}$, we must use a uniform grid over the input space $\mathbf{\Theta}$ in order to cook meals from diverse configurations. For the evaluation function $y = f(\mathbf{x}) \in [0,10]$, we have chosen a jury of $N$ experts to provide an evaluation $y_i$ for every recipe. The final evaluation $y = f(\mathbf{x})$ of every recipe is given by the mean of these experts $y = \frac{\sum_{i=1}{N} y_i}{N}$.

In this process, the chefs will acquire expert knowledge $\mathbf{\Phi}$ about the different variables $\mathbf{\phi}$ involved in every different recipe and how they link to the quality $y$. We propose to encode this expert knowledge $\mathbf{\Phi}$ into a set of conditional probability density functions $\mathbf{\Xi}$ to augment the dataset $\mathbf{D}_{real}$ by sampling the quality $y$ from them. For example, we know that if the temperature of the oven is very high, the quality of the recipe will be low with high security, we can model this expert knowledge by a Gamma Distribution $y \sim \Gamma(\cdot | \theta_1)$. Other distributions such as Gaussian have also been used, to model for example that a particular cooking time frame generated good quality meals $y \sim \mathcal{N}(\cdot | \theta_2)$. We fit the parameters of these conditional distributions $\mathbf{\Xi}$ by real data or by expert knowledge. By sampling points from these distributions $\mathbf{\Xi}$, we provide the latent noise of the objective function $y(\theta) = f(\theta) + \epsilon : \epsilon \sim \mathcal{N}(\mu, \sigma)$. We sample points from this set of distributions $\mathbf{\Xi}$ to build the dataset fitted by the ML model $\mathcal{D} = \mathcal{D}_{sim} \cup \mathcal{D}_{real}$.

We now fit an ML model $m$ and its hyperparameters $\mathbf{\rho}$ to this data, choosing the best ML model $m = \arg \min v(\mathbf{M})$. In order to do so, we propose to use 10-fold cross validation $v(m)$ over a pipeline of ML models $\mathbf{M}$. We can use BO, random or grid search to find the best model. The approximation to the jury of experts evaluation of the taste of the recipes will be given by the prediction of the chosen ML algorithm $m(\mathbf{\theta}) \approx f(\mathbf{\theta})$. Finally, we use BO having as objective function the prediction of the fitted ML model $m(\mathbf{\theta})$ of the recipes $\mathbf{\theta}$, avoiding the use of statistical tests as BO is an efficient search mechanism in this scenario. The suggested recipe $\mathbf{\theta}^*$ will be given by the best observation found by BO in a certain number of iterations. By following this methodology, we obtain a way to optimize a recipe and suggest a prototype without human intervention nor the need to cook in the optimization process. 

\section{Experiments}

We have performed two experiments to show the results of our proposed methodology. More experiments can be done following these steps, but these two combine real, integer and categorical variables, Table 1, resulting in a complete coverage of the problem. The first experiment involves the cooking process of a Hot Dog and the second one of a Cesar Salad. For both problems, $45$ different meals have been cooked and tested by experts. Then, we have codified the gained expert knowledge into probability distributions, where we have generated more instances.

\begin{table}[h!]
\centering
\begin{tabular}{ |p{4cm}||p{2.5cm}|p{4.2cm}||p{1.5cm}|  }
 \hline
 \multicolumn{2}{|c|}{Cesar Salad} & 
 \multicolumn{2}{|c|}{Hot Dog} \\
 \hline
 Variable Name & Bounds & Variable Name & Bounds \\
 \hline
 Sausage cooking time (s.)   & [0,200]    & Ceramic intensity (bread)&   [1,9]\\
 Bread cooking time (s.) &   [0,200]  & Bread cooking time (s.)  & [5,50]\\
 Cooking place & [pan,microwave] & Ceramic intensity (chicken) &  [1,9] \\
 Bbq sauce teaspoons & [0,9] & Chicken cooking time (s.) &  [1,200]\\
 Mayonnaise teaspoons & [0,9]  & Cesar Brand & [X,Y]\\
 Mustard teaspoons & [0,9]  & Lettuce Brand   & [X,Y,Z,T]\\
 \hline
\end{tabular}
\caption{Input space of the performed experiments.}
\end{table}

We have fixed an SVR model for the generated data. In order to find the best values of the hyperparameters of this model that generate the lowest medium squared error of a $10$ fold cross-validation over the generated dataset, we have used a grid search, testing different values for $C$  and $\gamma$. For both experiments, the optimum hyperparameters have been $C=1$ and $\gamma = 0.01$. We obtain errors of $2.56$ and $2.6$ points, which are very reasonable, since the probability distributions add noise to the evaluation function to model the subjectivity of this function.

Having fixed the prediction model, we know are going to test our hypothesis that states that using BO we will improve the quality of the recipe given by an expert. We predict the quality of the expert criterion using the model for both experiments. We also compare the BO optimization with a Random Search (RS). For BO, we have used a M\'atern Kernel for the GP. We optimize the hyperparameters of the GP by maximizing their log-marginal likelihood. We use 10 GP samples to build an averaged PES acquisition function. In Figure 1, we plot the quality of the recipe in every iteration of the process. We show how BO outperforms the expert and random search criterion easily, which is an empirical proof of the good behaviour of BO in these scenarios.

\begin{figure}[htb]
        \begin{center}
	\begin{tabular}{cc}
        \includegraphics[width=0.45\linewidth]{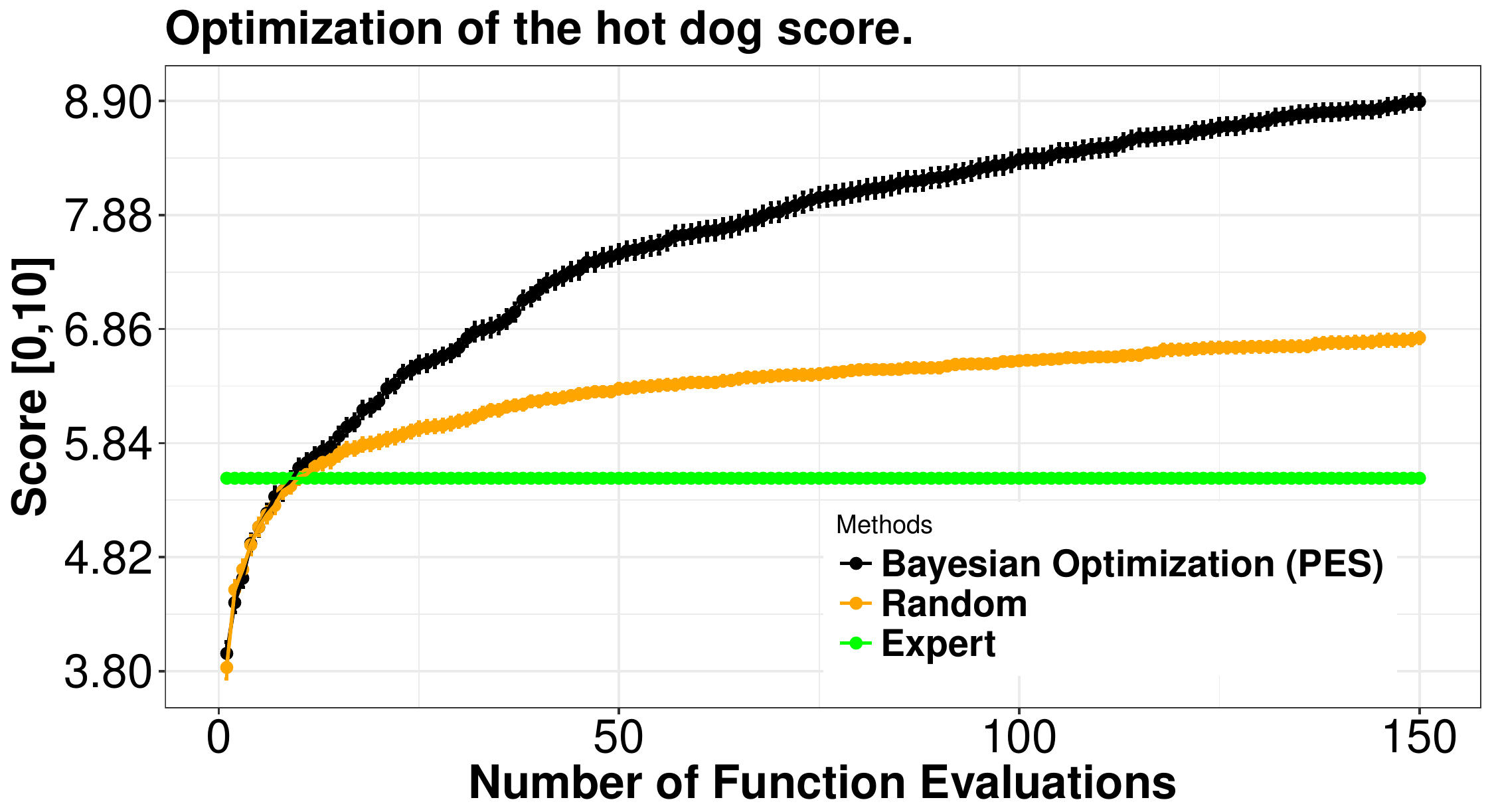}  &
        \includegraphics[width=0.45\linewidth]{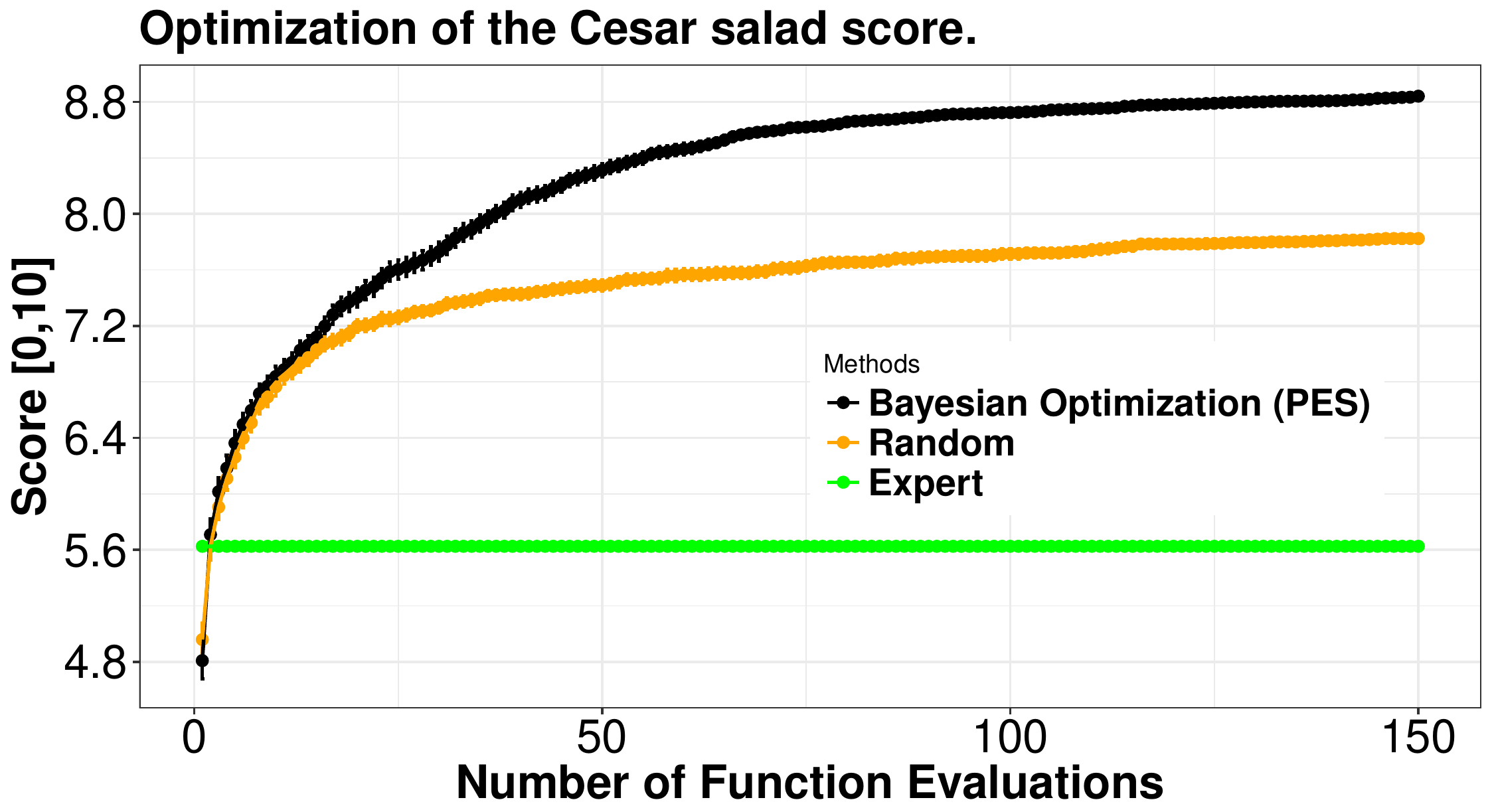} \\
	\end{tabular}
	\end{center}
	\vspace{-0.5cm}
	\caption{{\small Average results of the quality of the two real experiments by BO, RS and Expert Criterion approaches.}}
\end{figure}
\vspace{-0.5cm}
Figure 2 shows us histograms of the suggestions given by BO for all the 100 replications of the experiment. The suggested recipe given by the optimization methodology is formed by the union of the most suggested values of all the variables, shown in Table 2. The suggested values have been tested a posteriori and the experts agree on the quality of the meals given by BO.

\begin{figure}[htb]
        \begin{center}
        \begin{tabular}{cc}
        \includegraphics[width=0.45\linewidth]{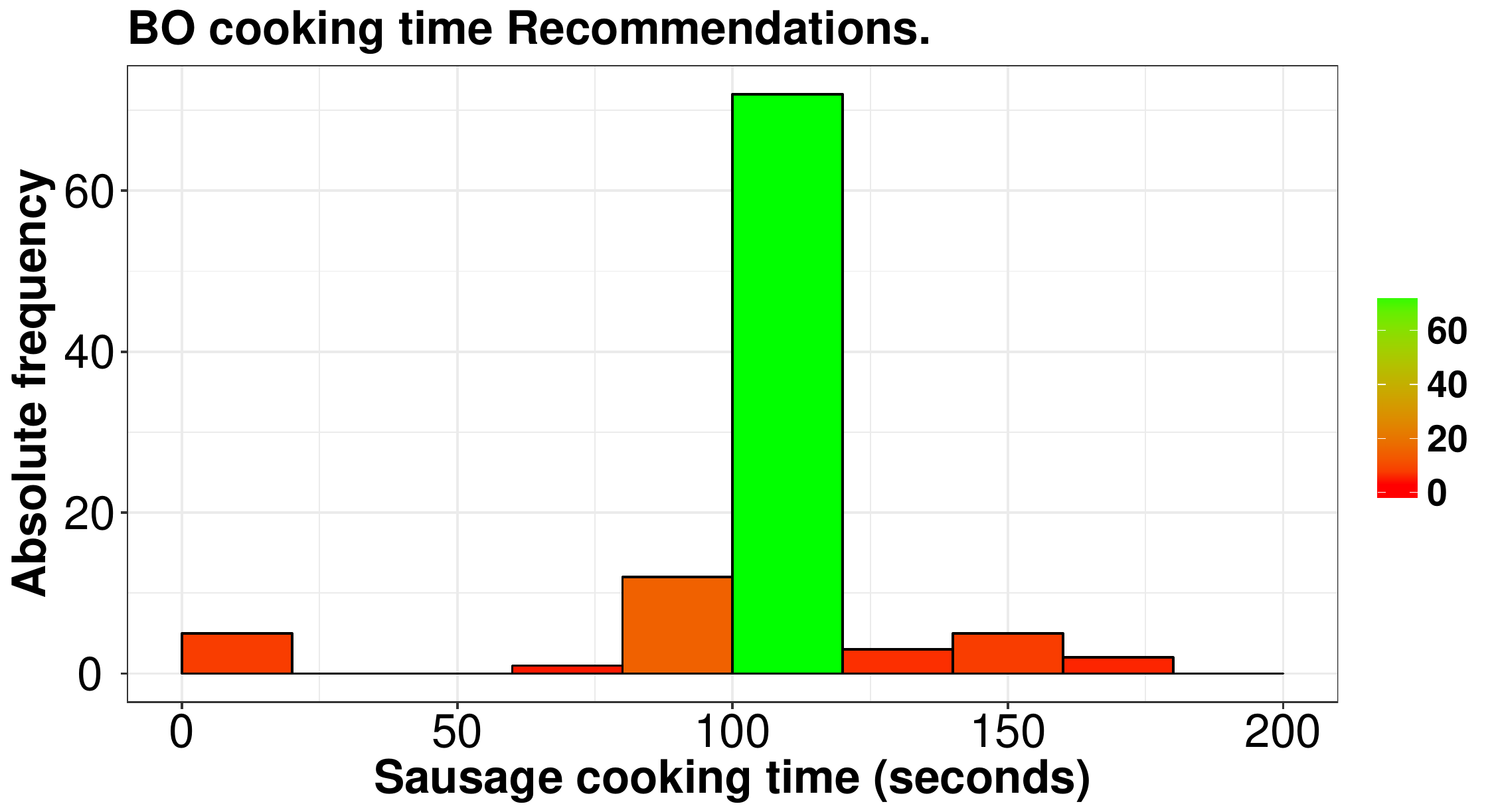}  &
        \includegraphics[width=0.45\linewidth]{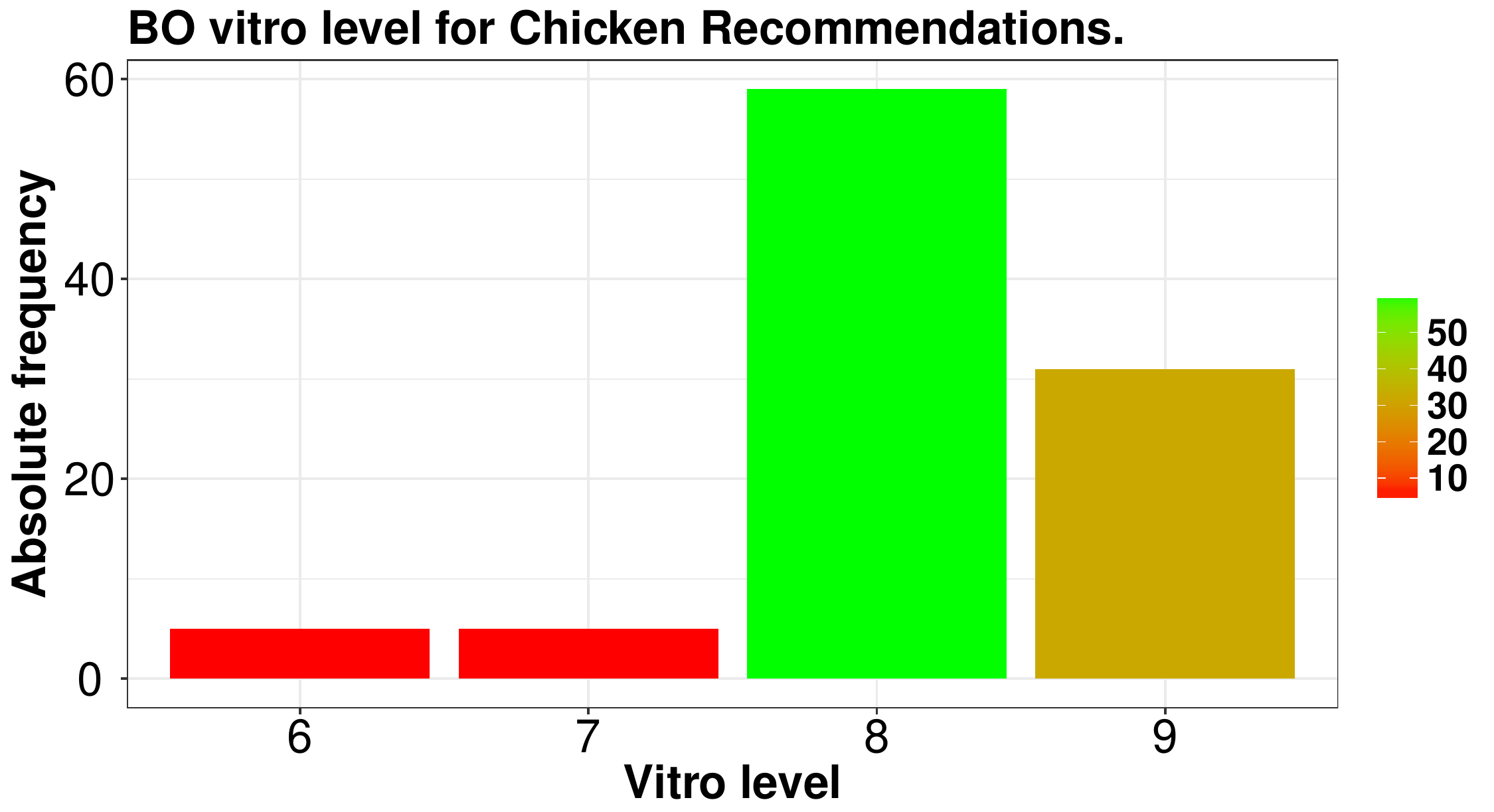} \\
        \end{tabular}
        \end{center}
        \caption{{\small Two histograms of proposed values for the variable of the two real experiments by the 100 replications of BO performed for the two experiments.}}
\end{figure}

\begin{table}[h!]
\centering
\begin{tabular}{ |p{4cm}||p{2.5cm}|p{4.2cm}||p{1.5cm}|  }
 \hline
 \multicolumn{2}{|c|}{Cesar Salad} &
 \multicolumn{2}{|c|}{Hot Dog} \\
 \hline
 Variable Name & Value & Variable Name & Value \\
 \hline
 Sausage cooking time (s.)   & [45,55]    & Ceramic intensity (bread)& 9\\
 Bread cooking time (s.) &   [100,115]  & Bread cooking time (s.)  & [12,15]\\
 Cooking place & pan & Ceramic intensity (chicken) &  8 \\
 Bbq sauce teaspoons & 0 & Chicken cooking time (s.) & [70,100]\\
 Mayonnaise teaspoons & 0  & Cesar Brand & X\\
 Mustard teaspoons & 0 & Lettuce Brand   & Y\\
 \hline
\end{tabular}
\caption{Recipes generated by the best observed result of BO. As the result is the most voted amongst 100 replications, bins have been created for integer and real-valued variables. Most voted bins are shown in the table.}
\end{table}
\section{Conclusions and Further Work}

We have proposed a methodology for generating cooking recipes from an input space of variables such as ingredients, tools and other variables. As we are restricted to our human and time resources we have simulated the process of cooking by an ML model, an SVR, that fits real and synthetic data generated by expert knowledge. The choices of the hyperparameters for BO, the ML model and the probability distributions mentioned in this methodology can be varied depending on the problem at hand. We can even use a metaheuristic such as a genetic algorithm or a particle swarm optimization to optimize instead of BO, but in a real case scenario, we will need BO due to the fact that the evaluations are too expensive both in time and human resources.

We have provided two real experiments that show the benefits of using BO to optimize the input space of cooking variables and compared it to a RS and an Expert Criterion. We replicate these experiments and provide the most voted recommendation by the BO approach. In both scenarios, BO outperforms the results obtained by the other approaches. Other objectives could have been modelled such as, for example, the presentation of the cooking recipe. Restrictions can also appear in the cooking process such as the time needed to elaborate a recipe or the total price of the recipe. These characteristics could be modelled by a Multi-Objective Constrained scenario that can be solved by Multi-Objective Constrained BO \citep{garrido2016predictive}. We are also planning to implement an interface that, given an input space and choices such as the optimization algorithm, the ML model and the expert knowledge rules it suggests a recipe written in natural language.

{\small
{\bf Acknowledgments}:
The authors acknowledge the use of the facilities of Centro
de Computaci\'on Cient\'ifica (CCC) at Universidad Aut\'onoma de
Madrid. The authors also acknowledge financial support from the Spanish
Plan Nacional I+D+i, Grants TIN2016-76406-P, TEC2016-81900-REDT
(MINECO/FEDER EU), and from Comunidad de Madrid, Grant S2013/ICE-2845.
}

\bibliographystyle{plainnat}
\bibliography{samplepaper}

\end{document}